\documentclass[sigconf]{acmart}

\usepackage{multicol,graphicx,mathrsfs,amsmath,pifont,amscd,latexsym,color,fancyhdr,CJK,url,longtable, pifont,subfig,epstopdf,setspace,multirow,colortbl,tabularx,threeparttable, booktabs, verbatim, bbm,listings, balance,color,indentfirst,xspace,hyperref,anyfontsize,mathtools}

\usepackage[linesnumbered,boxed,algosection]{algorithm2e}
\SetKwComment{tcp}{$\triangleright$ }{}

\SetCommentSty{mycommfont}

\usepackage{array}
\newcolumntype{L}[1]{>{\raggedright\let\newline\\\arraybackslash\hspace{0pt}}m{#1}}
\newcolumntype{C}[1]{>{\centering\let\newline\\\arraybackslash\hspace{0pt}}m{#1}}
\newcolumntype{R}[1]{>{\raggedleft\let\newline\\\arraybackslash\hspace{0pt}}m{#1}}

\newcommand{\xmark}{\ding{55}}%
\newcommand{\cmark}{\ding{51}}%
\newcommand{\sniffer}{DL Sniffer\xspace}
\newcommand{\extractor}{Model Extractor\xspace}

\newcommand{\modelnum}{176\xspace}

\setcopyright{none}

\acmConference[WWW'19]{The Web Conference}{May 2019}{San Francisco, USA}
\acmYear{2019}
\copyrightyear{2019}

\usepackage{wrapfig}
\usepackage{comment}
\usepackage{colortbl}

\usepackage[T1]{fontenc} 

\usepackage{color,soul}
\usepackage{graphics}
\usepackage{lipsum}

\usepackage{tikz}
\usetikzlibrary{decorations.pathmorphing, patterns,shapes,snakes} 

\usepackage[inline]{enumitem}	
\usepackage{listings} 
\usepackage{lipsum}

\usepackage{capt-of}	

\usepackage{ulem}  

\usepackage{mathrsfs}	
\usepackage{amsfonts}

\usepackage[export]{adjustbox} 

\definecolor{myForestGreen}{RGB}{34, 139, 34}
\definecolor{airforceblue}{rgb}{0.36, 0.54, 0.66}

\definecolor{refkey}{RGB}{34,139,34}
\definecolor{labelkey}{rgb}{0,0,1}

\renewcommand{\paragraph}[1]{\vskip 3pt\noindent\textbf{#1 }}	 

  {\begin{list}{$\bullet$}%
     {\setlength{\parsep}{0pt}%
      \setlength{\topsep}{0pt}%
      \setlength{\itemsep}{2pt}}}%
  {\end{list}}
\newcommand\noted[1]{} 

\newcommand{\textbfit}[1]{\textbf{\textit{#1}}}

\newenvironment{myitemize}%
  {\begin{itemize}
	[leftmargin=0cm,
		itemindent=.3cm,
		labelwidth=\itemindent,
		labelsep=0pt,
		parsep=3pt,
		topsep=2pt,
		itemsep=1pt,
		align=left]
  }%
  {\end{itemize}}    

\newenvironment{myenumerate}%
  {\begin{enumerate}
	[leftmargin=0cm,itemindent=.5cm,labelwidth=\itemindent,
		labelsep=0pt,
		parsep=1pt,
		topsep=1pt,
		itemsep=3pt,
		align=left]
  }%
  {\end{enumerate}}    
  
  {\begin{enumerate*}
	[label=\roman*)]
  }%
  {\end{enumerate*}}      

\makeatletter
\def\@copyrightspace{\relax}
\makeatother

\newcommand{\implication}[1]{{\vspace{0.15cm} \noindent \uline{\textbfit{Implications:}} \textit{#1}}}

\begin{document}
\title{A First Look at Deep Learning Apps on Smartphones}



\begin{abstract}
We are in the dawn of deep learning explosion for smartphones. 
To bridge the gap between research and practice, we present the first empirical study on 16,500 the most popular Android apps, demystifying how smartphone apps exploit deep learning in the wild. 
To this end, we build a new static tool that dissects apps and analyzes their deep learning functions. 
Our study answers threefold questions: what are the early adopter apps of deep learning, what do they use deep learning for, and how do their deep learning models look like. 
Our study has strong implications for app developers, smartphone vendors, and deep learning R\&D. 
On one hand, our findings paint a promising picture of deep learning for smartphones, showing the prosperity of mobile deep learning frameworks as well as the prosperity of apps building their cores atop deep learning. 
On the other hand, our findings urge optimizations on deep learning models deployed on smartphones, protection of these models, and validation of research ideas on these models.

\end{abstract}

 \author{Mengwei Xu, Jiawei Liu, Yuanqiang Liu}
 \affiliation{%
   \institution{Key Lab of High-Confidence Software Technologies (Peking University), MoE, Beijing, China}
   }
   \email{ {mwx, 1500012828, yuanqiangliu}@pku.edu.cn}


 \author{Felix Xiaozhu Lin}
 \affiliation{%
   \institution{Purdue ECE}
   \city{West Lafayette, Indiana, USA}}
   \email{xzl@purdue.edu}

 \author{Yunxin Liu}
 \affiliation{%
   \institution{Microsoft Research}
   \city{Beijing, China}}
\email{yunxin.liu@microsoft.com}
   
    \author{Xuanzhe Liu}\thanks{Xuanzhe Liu is the paper's corresponding author.}
 \affiliation{%
   \institution{Key Lab of High-Confidence Software Technologies (Peking University), MoE, Beijing, China}}
\email{xzl@pku.edu.cn}

%
%



\maketitle

\section{Introduction}

Being ubiquitous, smartphones are among the most promising platforms for deep learning (DL), the key impetus towards mobile intelligence in recent years~\cite{smartphoneai1,smartphoneai2,smartphoneai3,smartphoneai4}.
Such a huge market is driven by continuous advances in DL, including the introduction of latest neural network (NN) hardware~\cite{conf/asplos/ChenDSWWCT14,conf/fpga/ZhangLSGXC15,conf/isca/ChenES16,conf/isca/HanLMPPHD16}, improvement in DL algorithms~\cite{goodfellow2014generative,silver2017mastering,radford2015unsupervised,mnih2015human}, and increased penetration in huge information analytics~\cite{perozzi2014deepwalk,covington2016deep,okura2017embedding,grbovic2018real}.
The research community has built numerous DL-based novel apps~\cite{DBLP:conf/mobisys/ZengCZ17,conf/huc/BarzS16,conf/huc/MittalYGK16,conf/sensys/LaneBGFK15,conf/huc/RaduLBMMK16,conf/huc/LaneGQ15}.
The industry has also tried to utilize DL in their mobile products. For example, in the newly released Android 9 Pie OS, Google introduces a small feed-forward NN model to enable Smart Linkify, a useful API that adds clickable links when certain types of entities are detected in text~\cite{smartlinkify}.



Year 2017 marked the dawn of DL for smartphones.
Almost simultaneously, most major vendors roll out their DL frameworks for smartphones, or \textit{mobile DL framework} for short.
These frameworks include TensorFlow Lite (TFLite) from Google~\cite{tflite} (Nov. 2017), Caffe2 from Facebook~\cite{caffe2} (Apr. 2017), Core ML from Apple~\cite{coreml} (Jun. 2017), ncnn from Tencent~\cite{ncnn} (Jul. 2017), and MDL from Baidu~\cite{mdl} (Sep. 2017). 
These frameworks share the same goal: executing DL inference solely on smartphones. 
Compared to offloading DL inference from smartphones to the cloud~\cite{amazonml,googleml,azureml}, 
on-device DL inference better protects user privacy, 
continues to operate in the face of poor Internet connectivity, 
and relieves app authors from paying the expense of running DL in the cloud~\cite{conf/ipsn/LaneBGFJQK16,conf/asplos/KangHGRMMT17,conf/huc/LaneGQ15,MobileNet,conf/asplos/ChenDSWWCT14,conf/fpga/ZhangLSGXC15,conf/mobisys/LiuLZNLD18,conf/ipsn/LaneBGFJQK16,xu2018deepcache}. 


Following the DL framework explosion, there emerges the first wave of smartphone apps that embrace DL techniques. 
We deem it crucial to understand these apps and in particular how they use DL, because history has proven that such \textit{early adopters} heavily influence or even decide the evolution of new technologies~\cite{rogers2010diffusion} -- smartphone DL in our case.

To this end, we present the first empirical study on how real-world Android apps exploit DL techniques. 
Our study seeks to answer \textit{threefold} questions: 
what are the characteristics of apps that have adopted DL,
what do they use DL for, and
what are their DL models. 
Essentially, our study aims findings on how DL is being used by smartphone apps \textit{in the wild} and the entailed implications, 
filling an important gap between mobile DL research and practice. 


For the study, we have examined a large set of Android apps from the official Google Play market. 
We take two snapshots of the app market in early Jun. 2018 and early Sep. 2018 (3 months apart), respectively.
Each snapshot consists of 16,500 the most popular apps covering 33 different categories listed on Google Play. 
We generate insights by inspecting individual apps as well as by comparing the two snapshots. 
To automate the analysis of numerous Android apps, we build a new analyzer that inspects app installation packages, identifies the apps that use DL (dubbed ``DL apps''), and extracts DL models from these apps for inspection. 
To realize such a tool, we eschew from looking for specific code pattern and instead identify the usage of known DL frameworks, based on a rationale that most DL apps are developed atop DL frameworks.


Our key findings are summarized as follows. 

\paragraph{Early adopters are top apps}
(\S\ref{sec:app_characteristics})
We have found 211 DL apps in the set of apps collected in Sep. 2018. 
Only 1.3\% of all the apps, these DL apps collectively contribute 11.9\% of total downloads of all the apps and 10.5\% of total reviews. 
In the month of Sep. 2018, the 221 DL apps are downloaded for around 13,000,000 times and receive 9,600,000 reviews.
DL apps grow fast, showing a 27\% increase in their numbers over the 3 months in our study.

\paragraph{DL is used as core building blocks}
(\S\ref{sec:app_usage})
We find that 81\% DL apps use DL to support their core functionalities. 
That is, these apps would fail to operate without their use of DL. 
The number of such DL apps grow by 23\% over the period of 3 months. 


\paragraph{Photo beauty is the top use} (\S\ref{sec:app_usage})
DL is known for its diverse applications, as confirmed by the usage discovered by us, e.g. emoji prediction and speech recognition. 
Among them, photo beauty is the most popular use case: 94 (44.5\%) DL apps use DL for photo beauty; 61 (29\%) DL apps come from the photography category.


\paragraph{Mobile DL frameworks are gaining traction}
(\S\ref{sec:app_frameworks})
While full-fledged DL frameworks such as TensorFlow are still popular among DL apps due to their momentum, DL frameworks designed and optimized for constrained resources are increasingly popular. 
For instance, the number of DL apps using TFLite has grown by 258\% over the period of 3 months.

\paragraph{Most DL models miss obvious optimizations.}
(\S\ref{sec:model_structures})
Despite well-known optimizations, e.g. quantization which can reduce DL cost by up to two orders of magnitude with little accuracy loss~\cite{han2015deep}, 
we find only 6\% of DL models coming with such optimizations.

\paragraph{On-device DL is lighter than one may expect.} 
(\S\ref{sec:model_performance})
Despite the common belief that the power of DL models comes from rich parameters and deep layers, we find that DL models used in apps are very small, with median memory usage of 2.47 MB and inference computation of 10M FLOPs, which typically incurs inference delay of tens of milliseconds.
These models are not only much lighter than full models for servers (e.g. ResNet-50 with 200 MB memory and 4G FLOPs inference computations) but also lighter than well-known models specifically crafted for smartphones, e.g. MobileNet with 54 MB memory and 500M FLOPs inference computations. 

\paragraph{DL models are poorly protected.}
(\S\ref{sec:model_security})
We find only 39.2\% discovered models are obfuscated and 19.2\% models are encrypted. 
The remaining models are trivial to extract and therefore subject to unauthorized reuse.

\vspace{0.15cm}
\noindent
\textit{\textbf{Summary of implications:}}
Overall, our findings paint a promising picture of DL on smartphones, motivating future research and development. 
Specifically, the findings show strong implications for multiple stakeholders of the mobile DL ecosystem. 
\textbf{To app developers:}
our findings show that DL can be very affordable on smartphones;
developers, especially individuals or small companies, should have more confidence in deploying DL in their apps;
interested developers should consider building DL capability atop mobile DL frameworks; 
a few app categories, notably photography, are most likely to benefit from DL techniques. 
\textbf{To DL framework developers:} our findings encourage continuous development of frameworks optimized for smartphones; 
our findings also show the urgent need for model protection as the first-class concern of frameworks.
\textbf{To hardware designers:}
our findings motivate DL accelerator designs to give priority to the layers popular among mobile DL models. 
\textbf{To DL researchers:} our findings suggest that new proposal for optimizing DL inference should be validated on lightweight models that see extensive deployment on smartphones in the wild.

In summary, our contributions are as follows.
\begin{itemize}
\item We design and implement a tool for analyzing the DL adoption in Android apps. 
Capable of identifying the DL usage in Android apps and extracting the corresponding DL models for inspection, our tool enables automatic analysis of numerous apps for DL.

\item We carry out the first large-scale study of 16,500 Android apps for their DL adoption. 
Through the empirical analysis, we contribute new findings on the first wave of apps that adopt DL techniques. 
In the dawn of DL explosion for smartphones, our findings generate valuable implications to key stakeholders of the mobile ecosystem and shed light on the evolution of DL for smartphones.
We plan to publicize our tools and datasets. 
\end{itemize}

The remainder of the paper is organized as follows.
We describe the background knowledge and our motivations in Section~\ref{sec:back}.
We present our research goal and the analyzing tool which helps us identify DL usage and extract DL models in Section~\ref{sec:methodology}.
We present the analysis results of DL apps and DL models in Section~\ref{sec:app_analysis} and Section~\ref{sec:model_analysis}, respectively.
We discuss the limitations and possible future work in Section~\ref{sec:limitations}.
We survey the related work in Section~\ref{sec:related}, and conclude in Section~\ref{sec:conclusion}.

\section{Background}\label{sec:back}

\paragraph{DL models and frameworks}
DL has revolutionized many AI tasks, notably computer vision and natural language processing, through substantial boosts in algorithm accuracy. 
In practice, DL algorithms are deployed as two primary parts.
The first one is \textit{DL models}, which often comprise neuron layers of various types, e.g. convolution layers, pooling layers, and fully-connected layers.
Based on the constituting layers and their organizations, 
DL models fall into different categories, e.g., Convolutional Neural Network (CNN) containing convolution layers, 
and Recurrent Neural Network (RNN) processing sequential inputs with their recurrent sub-architectures. 
The second part is \textit{DL frameworks} that produce DL models (i.e. training) and execute the models over input data (i.e. inference).
Since a production DL framework often entails tremendous engineering efforts, most app developers tend to exploit existing frameworks by major vendors, such as \emph{TensorFlow} from Google.

\paragraph{Deploying mobile DL} 
As training models is intensive in both data and computing~\cite{li2014efficient}, 
smartphone developers often count on cloud servers for modeling training offline prior to app deployment.
At app installation time, 
the trained models are deployed as part of the app installation package. 
At runtime, apps perform inference with the trained models by invoking DL frameworks, and therefore execute AI tasks such as face recognition and language translation. 

\paragraph{Inference: on-cloud vs. on-device} 
Towards enabling DL on smartphones, model inference can be either offloaded to the cloud or executed solely on smartphones.
Offloading to the cloud is a classical use case of Software-as-a-Service (SaaS), and has been well studied in prior work~\cite{conf/imc/YaoXWVZZ17,journals/popets/HesamifardTGW18,conf/iml/BacciuCGM17,conf/icmla/RibeiroGC15}.
The mobile devices upload data and retrieve the inference results, transparently leveraging rich data center resources as server-class GPU.
Yet, we have observed on-device DL inference is quickly gaining popularity due to its unique advantages of stronger privacy protection, resilience against poor Internet connectivity, and lower cloud cost to app developers. 
We will present more evidence in the paper.
In this work, we focus our empirical study on such on-device deep learning for smartphones.

\section{Goal and Methodology}\label{sec:methodology}

\begin{figure*}[t]
	\centering
	\includegraphics[width=0.9\textwidth]{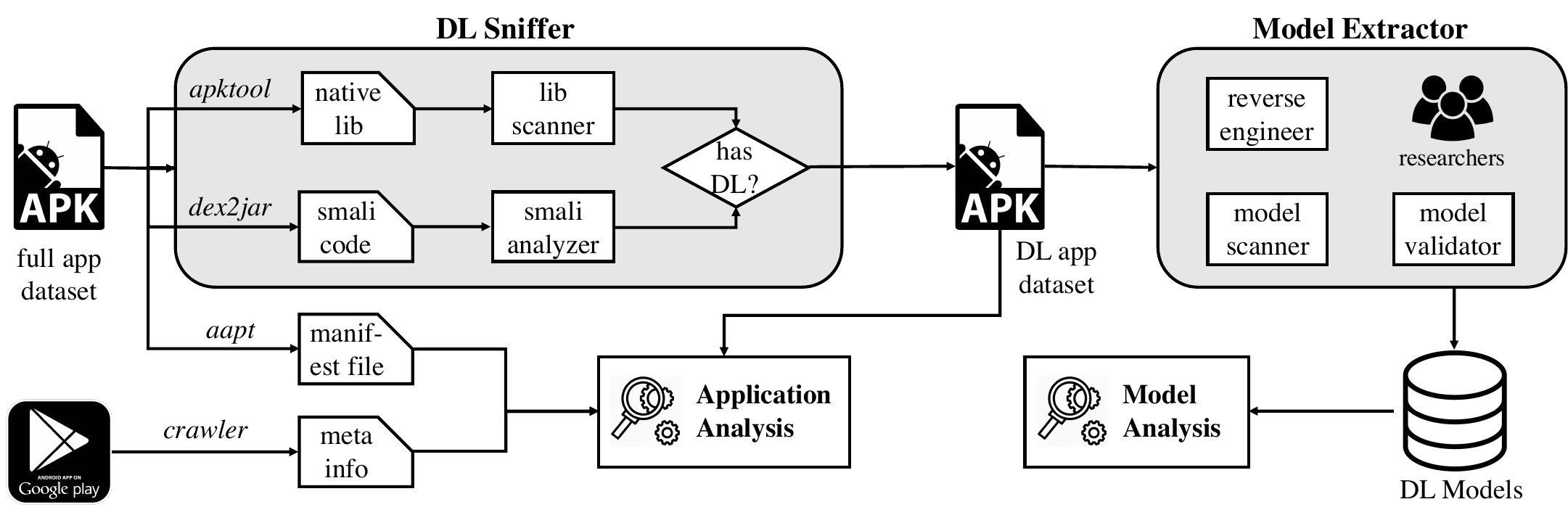}
	\caption{The overall workflow of our analyzing tool.}
	\label{fig:workflow}
\end{figure*}

\subsection{Research Goal}

The goal of our study is to demystify how smartphone apps exploit DL techniques in the wild. 
Our study focuses on two types of subjects: 
i) smartphone apps that embrace DL 
and ii) the DL frameworks and models used in practice. 
Accordingly, we characterize the apps, the frameworks, and the models. 
We will present the results in Section~\ref{sec:app_analysis} and Section~\ref{sec:model_analysis} respectively.

\paragraph{Scope}
We focus our analysis on Android apps, as Android represents
a dominant portion of smartphone shipment (88\% in the second quarter of 2018) and hence serves a good proxy for the entire smartphone app population~\cite{androidshare}.

\paragraph{Datasets} 
We retrieve from the Google Play store the full dataset used in this work.
We select 16,500 apps in total, which consist of 
the top 500 free apps with most downloads from each of the 33 categories defined by Google Play\footnote{Different app categories on Google Play can be visited via url \url{https://play.google.com/store/apps/category/XXX}, where XXX can be GAME or other category names.}.
We have crawled two datasets at different moments, June 2018 and September 2018, which are three months apart. 
The two app datasets have more than 2/3 overlapped apps. 
For each app, we download its apk file and crawl its meta information (e.g. app description and user rating) from the Google Play web page for analysis.
Our analysis primarily focuses on the newer dataset, i.e., Sep. 2018, and the difference between the two data sets, unless specified otherwise. 

\subsection{Workflow Overview}\label{sec:methodology_overview}

We design and implement an analyzing tool to enable our research goal on large-scale Android apps.
The tool runs in a semiautomatic way, as illustrated in Figure~\ref{fig:workflow}.

The very first step of the analyzing tool is identifying DL apps among a given set of Android apps as input. This is achieved via the module named \sniffer. The core idea of \sniffer, is detecting the usage of popular DL frameworks, instead of directly finding the usage of DL.
After identifying DL apps, it performs analysis on those apps. During analysis, we use the manifest files extracted from DL apps via tool aapt~\cite{aapt} and the meta information crawled from the corresponding Google Play web page. The manifest files include information such as package name, app version, required permissions, etc. The web pages include information such as app description, user rating, app developer, etc.

The analyzing tool further extracts DL models from those DL apps. This extraction is achieved via a module called \extractor.
After extracting DL models, it performs analysis on them. However, we here face the challenge that the models are mostly in different formats. Though developers are investing substantial efforts in integrating different model formats, such as designing a standardized one~\cite{onnx}, the ecosystem of DL frameworks is still broken and fragmented nowadays. Thus, when looking into the internal structures of DL models, we substantially leverage the available tools and source of different frameworks. Fortunately, most of the frameworks we investigated (details in Table~\ref{tab:overview_of_frameworks}) are open-source and well-documented.

We discuss more details of \sniffer and \extractor in Section~\ref{sec:sniffer} and Section~\ref{sec:extractor}, respectively.
\section{Application Analysis}\label{sec:app_analysis}
This section presents our analysis of smartphone DL apps. 
We first describe our methodology in Section~\ref{sec:sniffer} and then the following three major aspects of the DL apps: 

\begin{myitemize}
\item The characteristics of DL apps (\S\ref{sec:app_characteristics}): their popularity, their difference from non-DL apps, and their developers.

\item The role of DL (\S\ref{sec:app_usage}): 
the popular usage of DL in apps, the categories of DL apps, and evidence that DL is already used as core building blocks of apps.

\item An analysis of DL frameworks (\S\ref{sec:app_frameworks}): which frameworks are used, the cost, and their adoption trend over time.
\end{myitemize}

\subsection{Methodology: finding DL apps}
\label{sec:sniffer}
As a part of our analyzing tool (Section~\ref{sec:methodology_overview}),
\sniffer takes apk file(s) as input, and outputs which of them use DL technique. Detecting DL usage is difficult, instead \sniffer mainly detects the usage of popular DL frameworks with Android support.
Currently, \sniffer supports the detection of 16 popular DL frameworks, including \emph{TensorFlow, Caffe, TFLite,} etc, and the details of those frameworks will be presented later in Section~\ref{sec:app_frameworks} \& Table~\ref{tab:overview_of_frameworks}.
\sniffer uses two ways to mine the usage of DL frameworks: (1) For those who provide native C++ APIs, \sniffer first decomposes the apk files via Apktool~\cite{apktool}, and extracts the native shared libraries (with suffix ``.so''). \sniffer then searches for specific strings on the \textit{rodata} section of those libraries. Those strings can be regarded as identifications of corresponding frameworks, and are pre-defined by us. For example, we notice that the shared libraries that use \emph{TensorFlow} always have \textit{``TF\_AllocateTensor''} in its \textit{rodata} section. (2) For those who only support Java APIs, \sniffer first converts the apk file into smali code via dex2jar~\cite{dex2jar}. The smali code, which is a disassembled version of the DEX binary used by Android's Davik VM, enables us to carry out static program analysis. \sniffer statically goes through the class/method names of smali code and checks whether certain APIs exist. For example, the class \emph{MultiLayerConfiguration} is used in almost every app that embeds DeepLearning4J framework.

Besides detecting the usage of DL frameworks, we also try to identify DL apps that don't use the frameworks listed in Table~\ref{tab:overview_of_frameworks} (called ``no lib'' in this work). Similarly, this is achieved by searching specific strings on the \textit{rodata} section of native libraries as mentioned above, but the strings we use here are pre-defined such as \textit{``neural network'', ``convolution'', ``lstm''}, etc, rather than extracted from existing frameworks. We then manually check the detection correctness through reverse engineering, filtering those don't really have DL usage (false positive). This manual check is also performed on other DL apps detected using the aforementioned approach, to ensure good accuracy in DL identification.

\subsection{Characteristics of DL apps}\label{sec:app_characteristics}

\noindent $\bullet$ \textbf{Is DL gaining popularity on smartphones?}
Our study shows: over our investigation period (June 2018 -- Sept. 2018),
the total number of DL apps has increased by 27.1\%, from 166 to 211. 
We further investigate the new downloads and reviews of DL apps within one month of Sept. 2018: 
in that period, the 221 DL apps are downloaded for around 13,000,000 times and receive 9,600,000 new reviews. 
The results indicate a substantial amount of smartphones are running DL apps nowadays.

\begin{figure}[t]
	\centering
	\includegraphics[width=0.48\textwidth]{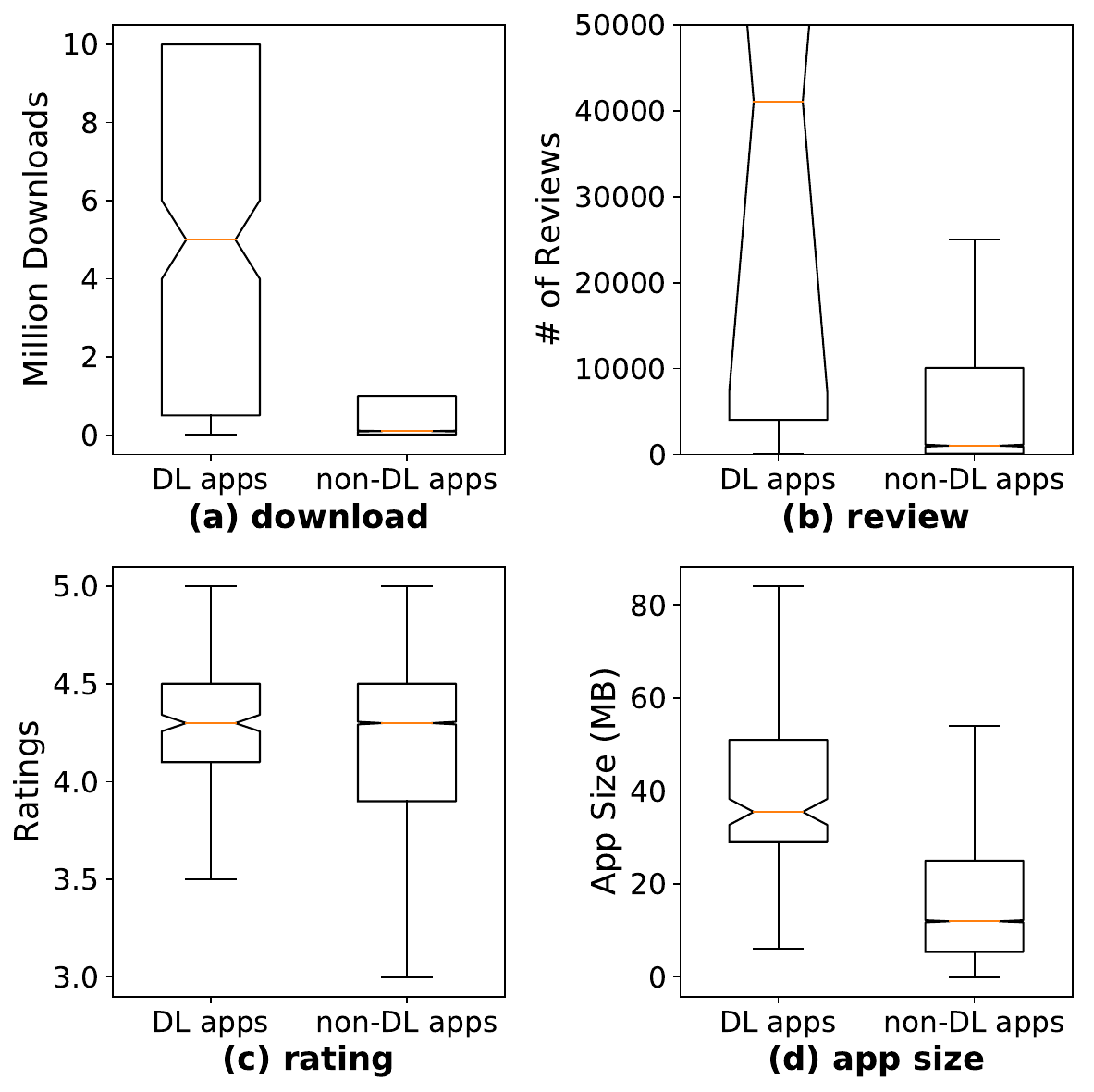}
	\caption{Comparisons between DL apps and non-DL apps on various aspects (a)--(d). 
	Each box shows the 75th percentile, median, and 25th percentile from top to bottom.
	We manually set the y-axis limits for better presentation in (b): the missed out 75th percentile of DL apps is 324,044.}
	\label{fig:compare}
\end{figure}

\noindent $\bullet$ \textbf{How are DL apps different from non-DL apps?}
We investigate the following three aspects with results illustrated in Figure~\ref{fig:compare}. 

\textit{Downloads and Reviews} are representative of apps' popularity. As observed, the median number of downloads and reviews of DL apps are 5,000,000 and 41,074 respectively, much larger than non-DL apps, i.e., 100,000 and 1,036 respectively.
We also count the download rankings of each DL apps within the corresponding category. The median number of such ranking is 89 among total 500 apps for each category.
We deem the above statistics as strong evidences that \textit{top apps are early adopters in deploying DL in mobile apps.} Such phenomenon can be explained that making DL work in the wild, though appealing, takes a lot of engineering efforts. The cycle of developing DL functionality on smartphones includes model construction, data collection, model training, offline/online testing, etc. Thus, many small companies or individual developers lack the resources to exploit DL on their apps.

\textit{Ratings} show how much appreciation users give to apps. The Figure shows that DL apps and non-DL apps have similar ratings from users, with the same median number 4.3.

\textit{App size}: as shown in Figure~\ref{fig:compare}, DL apps have much larger apk files than non-DL apps (median number: 35.5MB vs 12.1MB). This is reasonable since having DL not only adds DL frameworks and models to the apps, it also confidently indicates that the apps have much richer features.

\noindent $\bullet$ \textbf{Who are the developers of DL apps?}
We also study the developers of DL apps.
The results show that the identified 211 DL apps belong to 172 developers (companies), among which 27 developers have more than one DL apps. The developers with most DL apps are ``Google LLC'' (10) and ``Fotoable,Inc'' (6).
We observe many big companies own more than one DL apps, including Google, Adobe, Facebook, Kakao, Meitu, etc. This suggests that those big companies are pioneers in adopting DL into their products. We also notice that the DL apps from same developer often have identical DL frameworks. For example, four products from Fotoable Inc use the exactly same native library called \textit{libncnn\_style.0.2.so} to support DL technique. This is because that DL frameworks and even the DL models are easily reusable: a good nature of DL technique that can help reduce the engineering efforts of developers.

\implication{
The popularity of DL among top smartphone apps, especially ones developed by big companies, should endow smaller companies or independent developers with strong confidence in deploying DL in their apps.
}

\subsection{The roles of DL in apps}
\label{sec:app_usage}

\begin{table}[]
\begin{tabular}{|l|l|r|}
\hline
\rowcolor[HTML]{C0C0C0} 
\multicolumn{1}{|c|}{\cellcolor[HTML]{C0C0C0}\textbf{usage}} & \multicolumn{1}{c|}{\cellcolor[HTML]{C0C0C0}\textbf{detailed usage}} & \textbf{as core feature} \\ \hline
 & photo beauty: 97 & 94 (96.9\%) \\ \cline{2-3} 
 & face detection: 52 & 44 (84.6\%) \\ \cline{2-3} 
 & augmented reality: 19 & 5 (26.3\%) \\ \cline{2-3} 
 & face identification: 8 & 7 (87.5\%) \\ \cline{2-3} 
 & image classification: 11 & 6 (54.5\%) \\ \cline{2-3} 
 & object recognition: 10 & 9 (90\%) \\ \cline{2-3} 
\multirow{-7}{*}{image: 149} & text recognition:11 & 4 (36.3\%) \\ \hline
 & word\&emoji prediction: 15 & 15 (100\%) \\ \cline{2-3} 
 & auto-correct: 10 & 10 (100\%) \\ \cline{2-3} 
 & translation: 7 & 3 (42.8\%) \\ \cline{2-3} 
 & text classification: 4 & 2 (50\%) \\ \cline{2-3} 
\multirow{-5}{*}{text:26} & smart reply: 2 & 0 (0\%) \\ \hline
 & speech recognition: 18 & 7 (38.9\%) \\ \cline{2-3} 
\multirow{-2}{*}{audio: 24} & sound recognition: 8 & 8 (100\%) \\ \hline
 & recommendation: 11 & 2 (18.1\%) \\ \cline{2-3} 
 & movement tracking: 9 & 4 (44.4\%) \\ \cline{2-3} 
 & simulation: 4 & 4 (100\%) \\ \cline{2-3} 
 & abnormal detection: 4 & 4 (100\%) \\ \cline{2-3} 
 & video segment: 2 & 1 (50\%) \\ \cline{2-3} 
\multirow{-6}{*}{other: 19} & action detection: 2 & 0 (0\%) \\ \hline
\multicolumn{2}{|c|}{total: 211} & 171 (81.0\%) \\ \hline
\end{tabular}
\caption{The DL usage in different apps. Note: as one app may have multiple DL uses, the sum of detailed usage (column 2) might exceed the corresponding coarse usage (column 1).}
\label{tab:usage}
\end{table}

\noindent $\bullet$ 
\textbf{What are the popular uses of DL?}
To understand the roles played by DL, we manually classify the usage of DL on different apps. This is achieved by looking into the app description and app contents. The results are shown in Table~\ref{tab:usage}. Each app has one or more usages, and the usage is represented in two different levels (coarse and detailed). 10 apps are left out since we cannot confirm their DL usage.

Overall, image processing is the most popular usage of DL on smartphones, far more than text and audio processing (149 vs. 26 \& 24).
This is not surprising since computer vision is the field where DL starts the revolution~\cite{journals/corr/IandolaMAHDK16}, and the progress on this field has been lasting since then~\cite{lecun2015deep}.
In more details, photo beauty (97) and face detection (52) are mostly widely used in DL apps, usually found in photo editing and camera apps to beautify pictures. In text field, word \& emoji prediction (15) and auto-correct (10) are also popular, usually found in input method apps like GBoard. For audio processing, DL is mainly used for speech recognition (14). Besides, there are other types of usage such as recommendation (11) which is often found in shopping apps.

\begin{figure}[t]
	\centering
	\includegraphics[width=0.49\textwidth]{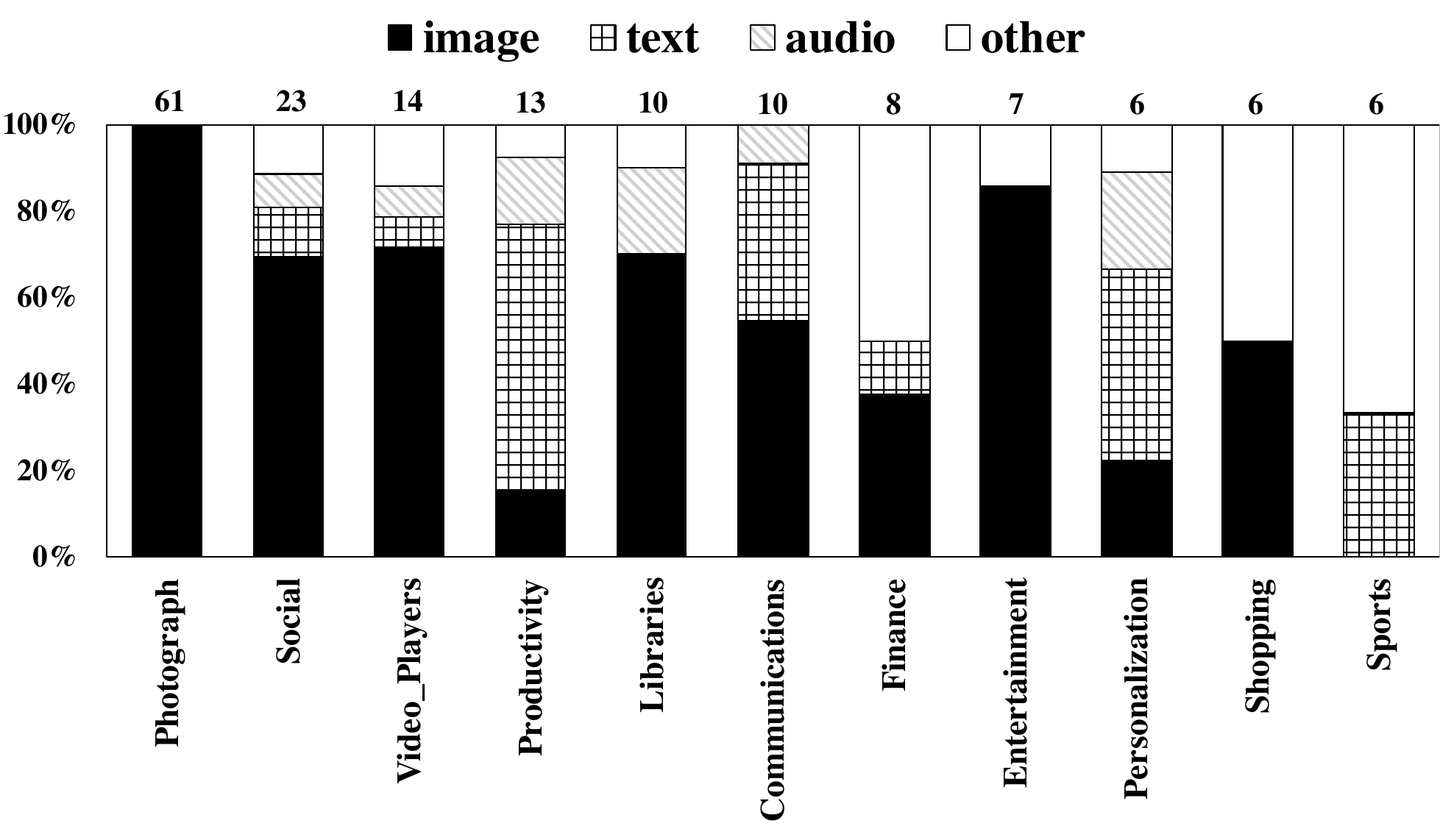}
	\caption{Distributions of DL apps over categories defined by Google Play. 	Numbers on top: the counts of DL apps in the corresponding categories. Apps in each category are further broken down by DL usage (see Table~\ref{tab:usage} for description). 
	Categories with fewer than 5 DL apps are not shown. 
}
	\label{fig:number_of_categories}
\end{figure}

\noindent $\bullet$ \textbf{Which categories do DL apps come from?}
Figure~\ref{fig:number_of_categories} summarizes the number of DL apps in different categories. As observed, almost 29\% DL apps (61 out of 211) are in category photograph, all of which use DL for image processing.
Social category is another hotspot with 23 DL apps in total, 78\% of which use DL for image processing while others use it for text, audio, etc.
The category of productivity also contains 13 DL apps, but most of them (62\%) use DL for text processing.
Overall, we can see that the DL usage is somehow diverse, with 11 categories has more than 5 DL apps among the top 500.
Such diversity gives credits to the good generality of DL technique.

\implication{
Our findings encourage developers of certain types of apps, notably the ones with photo beauty, to embrace DL.
Our findings also motivate encapsulating DL algorithms within higher level abstractions that cater to popular uses in apps. 
For instance, companies such as SenseTime already starts to ship DL-based face detection libraries to app developers.
Masking the details of DL models and frameworks, such abstractions would make DL more friendly to developers.}

\noindent $\bullet$ 
\textbf{Is DL a core building block?}
We also manually tag each DL usage as core feature or not. We define the DL functionality as apps' core feature if and only if two conditions are satisfied: (1) \textit{hot}: the DL functionality is very likely to be invoked every time the apps are opened and used by users, (2) \textit{essential}: without the DL functionality, the apps' main functionality will be severely compromised or even become infeasible. For example, DL on \textit{text recognition} is treated as core feature in a scanner app (Adobe Scan) that helps users translate an image into text, but not in a payment app (Alipay) that uses it to scan ID card for identification.
Similarly, DL on \textit{photo beauty} is treated as core feature in a camera app (Meitu), but not in a social app (Facebook Messenger Kids).

Overall, 171 out of 211 (81\%) apps use DL to support core features.
Specifically, since photo beauty (96.9\%) and face detection (84.6\%) are primarily used in photo \& camera apps, their usage is essential. Similarly, word \& emoji prediction (100\%) and auto-correct (100\%) are treated as core features in keyboard apps, helping users input more efficiently and accurately.
However, recommendation (18.1\%) is often provided as complementary feature to others such as shopping apps, thus not treated as core feature.

\implication{Our findings support future investment on R\&D of mobile DL, as core user experience on smartphones will likely depend on DL performance~\cite{conf/ipsn/LaneBGFJQK16} and security~\cite{papernot2016limitations,stoica2017berkeley}.}

\subsection{DL frameworks}\label{sec:app_frameworks}
As mentioned in Section~\ref{sec:back}, DL frameworks are critical to DL adoption, as most developers use those frameworks to build their DL apps. In this subsection, we investigate into how those frameworks are used in DL apps: the numbers, the sizes, the practice, etc.

\begin{figure}[t]
	\centering
	\includegraphics[width=0.48\textwidth]{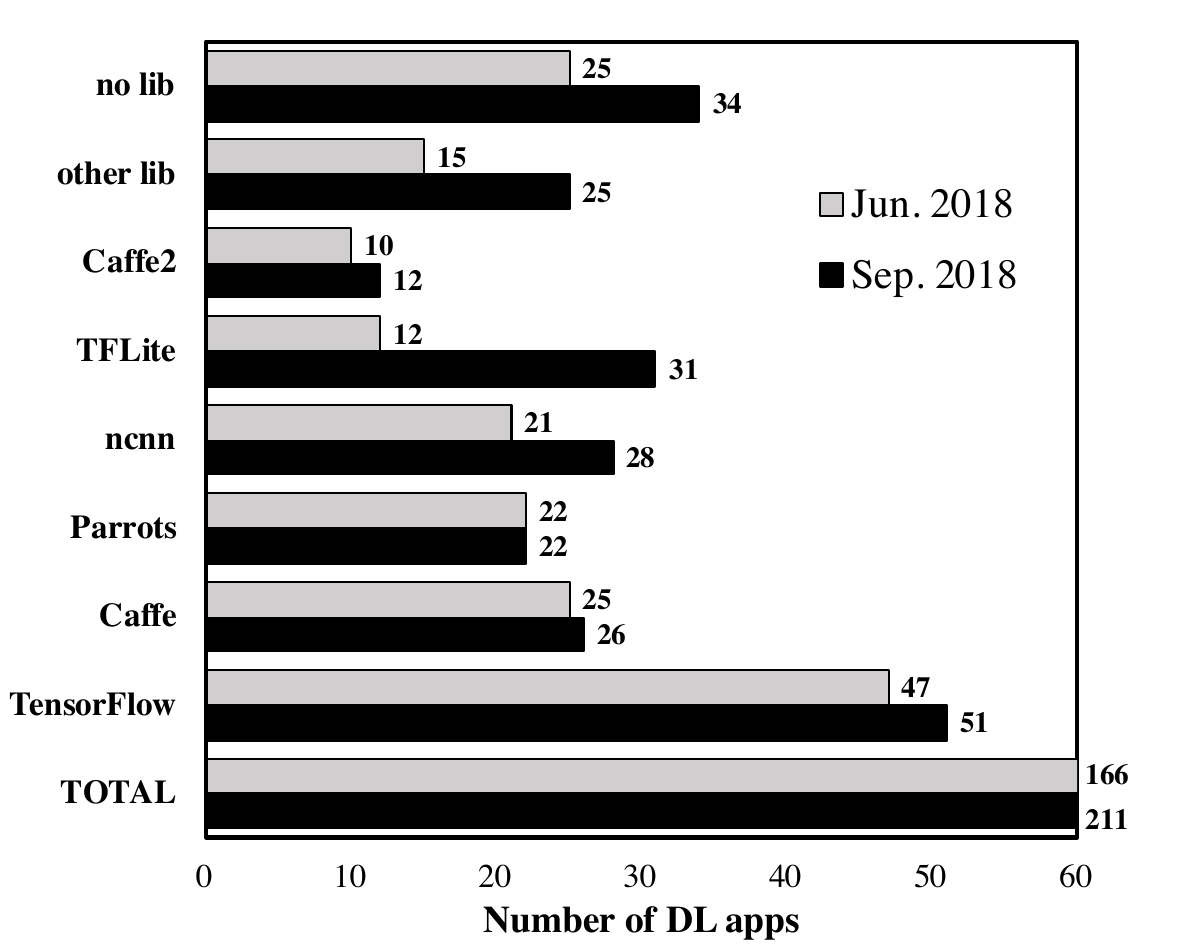}
	\caption{Numbers of DL apps using various mobile DL frameworks. 
	``other lib'': the DL apps developed on the frameworks in Table~\ref{tab:overview_of_frameworks} but not itemized here, e.g. \emph{mace, SNPE, and xnn}. 
	``no lib'': apps with DL functions but using no DL frameworks from Table~\ref{tab:overview_of_frameworks}. Note that the number in ``TOTAL'' is lower than the sum of others since some DL apps have integrated multiple frameworks.}
	\label{fig:number_of_frameworks}
\end{figure}

\begin{table*}[t]
\centering
\small
\begin{tabular}{|l|c|c|c|C{1.2cm}|C{3cm}|C{1.3cm}|}
\hline
\textbf{Framework} & \textbf{Owner} & \textbf{Supported Mobile Platform} & \textbf{Mobile API} & \textbf{Is Open-source} & \textbf{Supported Model Format} & \textbf{Support Training} \\ \hline
TensorFlow~\cite{TensorFlow} & Google & Android CPU, iOS CPU & Java, C++ & \cmark & ProtoBuf (\emph{.pb, .pbtxt}) & \cmark\\ \hline
TF Lite~\cite{tflite} & Google & Android CPU, iOS CPU & Java, C++ & \cmark & FlatBuffers (\emph{.tflite}) & \xmark\\ \hline
Caffe~\cite{caffe} & Berkeley & Android CPU, iOS CPU & C++ & \cmark & customized, json (\emph{.caffemodel, .prototxt}) & \cmark\\ \hline
Caffe2~\cite{caffe2} & Facebook & Android CPU, iOS CPU & C++ & \cmark & ProtoBuf (\emph{.pb}) & \cmark\\ \hline
MxNet~\cite{mxnet} & Apache Incubator & Android CPU, iOS CPU & C++ & \cmark & customized, json (\emph{.json, .params}) & \cmark\\ \hline
DeepLearning4J~\cite{dl4j}  & Skymind & Android CPU & Java & \cmark & customized (\emph{.zip}) & \cmark\\ \hline
ncnn~\cite{ncnn} & Tencent & Android CPU, iOS CPU & C++ & \cmark & customized (\emph{.params, .bin}) & \xmark\\ \hline
OpenCV~\cite{opencv} & OpenCV Team & Android CPU, iOS CPU & C++ & \cmark & TesnorFlow, Caffe, etc & \xmark\\ \hline
FeatherCNN~\cite{feathercnn} & Tencent & Android CPU, iOS CPU & C++ & \cmark & customized (\emph{.feathermodel}) & \xmark\\ \hline
PaddlePaddle~\cite{mdl} & Baidu & Android CPU, iOS CPU \& GPU & C++ & \cmark & customized (\emph{.tar}) & \cmark\\ \hline
xNN~\cite{xnn} & Alibaba & Android CPU, iOS CPU & unknown & \xmark & unknown & unknown\\ \hline
superid~\cite{superid} & SuperID & Android CPU, iOS CPU & unknown & \xmark & unknown & unknown\\ \hline
Parrots~\cite{sensetime} & SenseTime & Android CPU, iOS CPU & unknown & \xmark & unknown & unknown\\ \hline
MACE~\cite{mace} & XiaoMi & Android CPU, GPU, DSP & C++ & \cmark & customized (\emph{.pb, .yml, .a}) & \xmark\\ \hline
SNPE~\cite{snpe} & Qualcomm & Qualcomm CPU, GPU, DSP & Java, C++ & \xmark & customized (\emph{.dlc}) & \xmark\\ \hline
CNNDroid~\cite{cnndroid} & Oskouei et al. & Android CPU \& GPU & Java & \cmark & MessagePack (\emph{.model}) & \xmark\\ \hline
CoreML~\cite{coreml} & Apple & iOS CPU, GPU & Swift, OC & \xmark & customized, ProtoBuf (\emph{.proto, .mlmodel}) & \cmark\\ \hline
Chainer~\cite{chainer} & Preferred Networks & / & / & \cmark & customized (\emph{.chainermodel}) & \cmark\\ \hline
CNTK~\cite{cntk} & Microsoft & / & / & \cmark & ProtoBuf (\emph{.proto}) & \cmark\\ \hline
Torch~\cite{torch} & Facebook & / & / & \cmark & customized (\emph{.dat}) & \cmark\\ \hline
PyTorch~\cite{pytorch} & Facebook & / & / & \cmark & customized, pickle (\emph{.pkl}) & \cmark\\ \hline
\end{tabular}
\caption{An overview of popular deep learning frameworks and their smartphone support at the time of writing (Nov. 2018).}
\label{tab:overview_of_frameworks}
\end{table*}

\noindent $\bullet$ \textbf{A glance over popular DL frameworks} We first make an investigation into popular DL frameworks, and the results are summarized in Table~\ref{tab:overview_of_frameworks}. We select those 21 frameworks for their popularity, e.g., forks and stars on GitHub, gained attention on StackOverflow and other Internet channels. Among those 21 frameworks, 16 frameworks support Android platform via Java (official language on Android) and/or C++ (native support via cross-compilation). Most of them are open-source, while others are either provided to public as a binary SDK (\emph{SNPE, CoreML}), or only accessible by the providers' (collaborators') own products (\emph{xNN, Parrots}). Most of them use customized format to store and represent the model files, but some leverage existing approaches, such as ProtoBuf~\cite{protobuf}. We also notice a trend on lightweight DL inference frameworks, which are designed specifically for mobile apps but have no training-support back-end (\emph{ncnn, FeatherCNN, MACE}, etc). Those frameworks cannot train DL models, but can predict with pre-trained models via other frameworks such as \emph{TensorFlow} or \emph{Caffe}.
Note that our later analysis substantially relies on the openness of DL frameworks: it enables us to use the existing tools to analyze or even visualize the DL models such as \emph{TensorFlow}, or build our own interpreting scripts to analyze them based on the open code such as \emph{ncnn}.

\noindent $\bullet$ \textbf{What is the adoption of DL frameworks?} 
As summarized in Figure~\ref{fig:number_of_frameworks}, the most popular DL frameworks used in Sep. 2018 are \emph{TensorFlow} (51), \emph{TFLite} (31), and \emph{ncnn} (28), as they contribute to almost 50\% of the total number of DL apps. Other popular frameworks include \emph{Caffe}, \emph{Parrots}, and \emph{Caffe2}.
We have made several observations from those 6 dominant frameworks as following.
\begin{myenumerate}
\item 
All these frameworks are developed by big companies (e.g. Google), AI unicorns (e.g. SenseTime), or renowned universities (e.g. Berkeley).
\item
5 out of these 6 frameworks are open-source, except \emph{Parrots} which is provided as SDK to consumers.
In fact, it is believed that openness is already an important feature in machine learning, especially DL society, as it supposes to~\cite{sonnenburg2007need}.
It helps developers reproduce the state-of-the-art scientific algorithms, customize for personal usage, etc.
As a result, for example, \emph{TensorFlow} has more than 1,670 contributors up to Oct. 2018, going far beyond the community of Google.
\item
Most (4 out of 6) frameworks are optimized for smartphones, except \emph{Caffe} and \emph{TensorFlow}. Those mobile DL frameworks are designed and developed specifically for mobile devices, usually without training back-end, so that the resulted libraries can be faster and more lightweight.
As an example, \emph{TFLite} stems from \emph{TensorFlow}, but is designed for edge devices and reported to have lower inference time and smaller library size than \emph{TensorFlow}~\cite{zhang2018pcamp}.
Besides those popular frameworks, we identify 34 (16.1\%) DL apps that don't use any framework in Table~\ref{tab:overview_of_frameworks}.
These apps use self-developed engines to support DL functionality.
\end{myenumerate}

\noindent $\bullet$ \textbf{Are mobile DL frameworks gaining traction?}
As shown in Figure~\ref{fig:number_of_frameworks}, DL frameworks optimized for smartphones, such as \textit{TFLite} and \textit{ncnn}, quickly gain popularity:
the number of \emph{TFLite}-based DL apps has increased from 12 to 31;
that of \emph{ncnn} increases from 21 to 28. 
We deem it as the trend of mobile DL ecosystem: \textit{To train a model offline, use large, mature, and generic frameworks that focus on developer friendliness and feature completeness. To deploy a model on edge devices, switch to mobile-oriented frameworks that focus on performance (inference latency, memory footprint, and library size).}

\begin{figure}[t]
	\centering
	\includegraphics[width=0.48\textwidth]{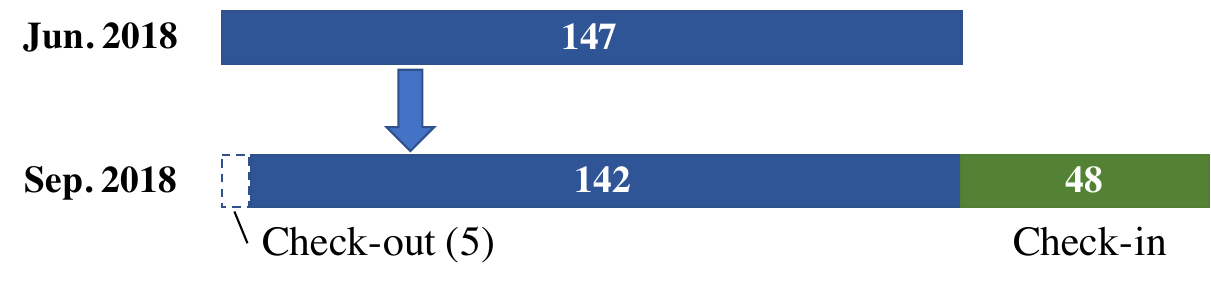}
	\caption{The number of check-in and check-out DL apps.}
	\label{fig:checkin}
\end{figure}

We also investigate into the DL check-in and check-out behavior in mobile apps. We define the check-in DL apps as those that have no DL usage in earlier version (Jun. 2018) but add the DL usage in newer version (Sep. 2018), and the check-out vice versa.
Note that the app list of our two datasets are not identical since we crawl the most popular ones, but the popularity is changing.
So we only consider the apps that exist in both lists (11,710 in total), and conclude the results in Figure~\ref{fig:checkin}.
As observed, 48 out of the 190 (25.3\%) DL apps in newer version are checked in between Jun. 2018 and Sep. 2018. We also notice that some DL apps in old version checked out, but the number is much smaller (5). The reasons of check-out can be that the developers remove the corresponding functionality or just replace the DL with other approaches.
Overall, the statistics support the fact that DL technique is increasingly adopted in mobile apps.

\begin{figure}[t]
	\centering
	\includegraphics[width=0.48\textwidth]{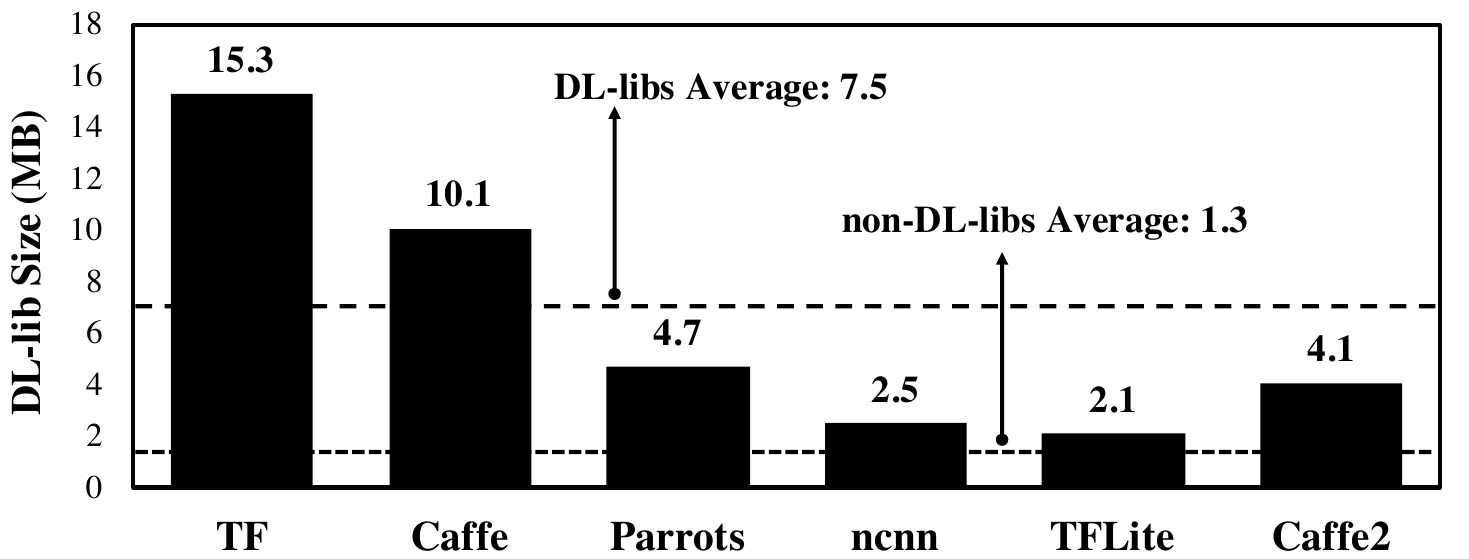}
	\caption{The binary library sizes of DL frameworks.}
	\label{fig:lib_size}
\end{figure}

\noindent $\bullet$ \textbf{What is the storage overhead of frameworks?}
Figure~\ref{fig:lib_size} shows the sizes of DL libs, i.e. the physical incarnation of DL frameworks. 
As shown, the average size of DL libs is 7.5MB, almost 6 times compared to the non-DL libs. Here, we only use the non-DL libs found within DL apps.
The results show that DL libs are commonly heavier than non-DL libs, because implementing DL functionality, even without training backend, is quite complex.
Looking into different frameworks, using \emph{TensorFlow} and \emph{Caffe} results in larger DL libs, i.e., 15.3MB and 10.1MB respectively, while others are all lower than 5MB. The reason is that mobile supports of \emph{TensorFlow} and \emph{Caffe} are ported from the original frameworks and substantially reuse the code base from them. However, these two frameworks are designed for distributed on-cloud DL. As comparison, other frameworks in Figure~\ref{fig:lib_size} are specifically designed for mobile devices to the purpose of good performance.

\paragraph{One app may incorporate multiple DL frameworks.}
Surprisingly, we find that 24 DL apps embed more than one DL frameworks. For example, AliPay, the most popular payment app in China, has both \emph{xnn} and \emph{ncnn} inside. We deem such multi-usage as (potentially) bad practice, since it unnecessarily increases the apk size and memory footprint when these frameworks need to be loaded simultaneously.
According to our statistics, the overhead is around 5.4MB, contributing to 13.6\% to the total apk size on average.
Such overhead can be avoided by running different tasks based on one framework, since most DL frameworks are quite generic and can support various types of DL models. Even if not, they can be easily extended to support the missing features~\cite{addop}.
The reason of such multi-usage behavior can be twofold. First, one app might be developed by different engineers (groups), who introduce different frameworks for their own DL purpose. Second, the developers may just reuse existing code and models for specific tasks, without merging them together in one DL implementation.

\implication{Our findings highlight the advantages and popularity of mobile DL frameworks, encouraging further optimizations on them.
Our findings also motivate app developers to give these frameworks priority considerations in choosing the incarnation of DL algorithms.
}
\section{Model Analysis}\label{sec:model_analysis}

This section focuses on the model-level analysis of DL technique.
We first describe the methodology details, e.g., the design of \extractor in Section~\ref{sec:extractor}.
Then, we show the analysis results on those DL models from three main aspects.
\begin{itemize}
\item The structures of DL models: the model types, layer types, and optimizations used (Section~\ref{sec:model_structures}).
\item The resource footprint of DL models: storage, memory, execution complexity, etc (Section~\ref{sec:model_performance}).
\item The security of DL models: using obfuscation and encryption to protect models from being stolen (Section~\ref{sec:model_security}).
\end{itemize}

\subsection{\extractor: finding DL models}\label{sec:extractor}
As a part of our analyzing tool (Section~\ref{sec:methodology_overview}), \extractor takes DL apps which we have already identified as input, and outputs the DL model(s) used in each app. \extractor scans the \emph{assets} folder of each decomposed DL apps, tries to validate each model file inside. Since DL frameworks use different formats to store their model files, \extractor has a validator for each of supported framework.
However, we observe that many models are not stored as plaintext inside apk files.
For example, some of them are encrypted on storage, and decrypted when apps running on devices.
For such cases, \extractor tries to reverse engineer the apps, and extract the analyzable models.

\textbf{Overall results} We extract DL models based on the most popular frameworks, i.e., \emph{TFLite}, \emph{TensorFlow}, \emph{ncnn}, \emph{Caffe}, or \emph{Caffe2}.
In summary, we successfully extract \modelnum DL models, which come from 71 DL apps.
The reasons why we cannot extract models from the other DL apps could be (i) the models are well protected and hidden in the apk files; (ii) the models are retrieved from Internet during runtime.
Among the extracted models, we can analyze 98 of them, which come from 42 DL apps.
The other extracted models cannot be parsed via our framework currently because (i) the models are in format which we have no insights into, such as \emph{Parrots}-based models, since the corresponding frameworks are not open-source; (ii) the models are encrypted.

\subsection{Model Structures}\label{sec:model_structures}
\begin{table}[]
\small
\begin{tabular}{|l|l|l||l|l|l|}
\hline
\textbf{\begin{tabular}[c]{@{}l@{}}Layer\\ type\end{tabular}} & \textbf{\begin{tabular}[c]{@{}l@{}}\% of \\ models\end{tabular}} & \textbf{\begin{tabular}[c]{@{}l@{}}\# in each\\ model\end{tabular}} & \textbf{\begin{tabular}[c]{@{}l@{}}Layer\\ type\end{tabular}} & \textbf{\begin{tabular}[c]{@{}l@{}}\% of \\ models\end{tabular}} & \textbf{\begin{tabular}[c]{@{}l@{}}\# in each\\ models\end{tabular}} \\ \hline
conv & 87.7 & 5 / 14.8 & relu & 51.0 & 6 / 16.3 \\ \hline
pooling & 76.5 & 2 / 2.8 & split & 46.9 & 1 / 7.5 \\ \hline
softmax & 69.1 & 1 / 1.1 & prelu & 32.1 & 4 / 4.6 \\ \hline
fc & 60.5 & 3 / 5.6 & reshape & 28.4 & 2 / 24.1 \\ \hline
add & 56.8 & 9.5 / 23.8 & dropout & 21.0 & 1 / 1.0 \\ \hline
\end{tabular}
\caption{Layers used in DL models. ``\% of models'' shows how many models contain such layer, while ``\# in each model'' shows the median/mean numbers of occurrences in each model that contains such layer.
``conv'' and ``fc'' are short for convolution and fully-connect.}
\label{tab:layers}
\end{table}


\noindent $\bullet$ \textbf{DL model types}
Among the DL models extracted, 87.7\% models are CNN models, 7.8\% models are RNN models, while others are not confirmed yet. The CNN models are mostly used in image/video processing and text classification. The RNN models are mostly used in text/voice processing, such as word prediction, translation, speech recognition, etc.
The results are consistent with the conventional wisdom: CNN models are good at capturing visual characteristics from images, while RNN models are powerful at processing sequential data with temporal relationships such as text and audio.

\noindent $\bullet$ \textbf{DL layer types}
We then characterize different types of layers used in DL models.
As shown in Table~\ref{tab:layers}, convolution (conv) is the most commonly used type. 87.7\% models have at least one convolutional layer, and the median (mean) number of convolutional layers used in those models is 5 (14.8). This is not surprising since convolution is the core of CNN models, and CNN is the dominant model architecture used in vision tasks. Our previous analysis already demonstrates that image processing is the most popular use case of mobile DL. Similarly, pooling is also an important layer type in CNN models, included in 76.5\% DL models. Besides, softmax is also frequently used (69.1\%), but don't repeatedly show up in one single model. This is because softmax layer usually resides at the end of DL models to get the probabilities as output.
As a comparison, fully-connected (fc) layers are less common, only used in 60.5\% DL models. A possible reason is that fully-connected layer is known to be very parameter- and computation-intensive, and can be replaced by other layers such as convolution~\cite{fcreplace}. Other frequently used layer types include add, split, relu, prelu, dropout, and reshape.

\implication{
Our findings motivate framework and hardware vendors who are interested in optimizing mobile DL to focus on the popular DL layers we discovered in deployed models, e.g. convolution.}

We also notice that a small number (5) of DL models contain customized layer types. Such customization is made as an extension to existing DL frameworks~\cite{addop}. The result indicates that the functionalities of current DL frameworks are mostly complete enough.

\begin{table}[t]
\small
\begin{tabular}{|l|l|l|l|l|}
\hline
& \textbf{1-bit Quan.} & \textbf{8-bit Quan.} & \textbf{16-bit Quan.} & \textbf{Sparsity}\\\hline
TF & unsupported & 4.78\% & 0.00\% & 0.00\%\\\hline
TFLite & unsupported & 66.67\% & unsupported & unsupported\\\hline
Caffe & unsupported & 0.00\% & unsupported & unsupported\\\hline
Caffe2 & unsupported & 0.00\% & unsupported & unsupported\\\hline
ncnn & unsupported & 0.00\% & unsupported & unsupported\\\hline
\textit{Total} & \textit{0.00\%} & \textit{6.32\%} & \textit{0.00\%} & \textit{0.00\%}\\\hline
\end{tabular}
\caption{Optimizations applied on DL models.}
\label{tab:model_opts}
\end{table}
\noindent $\bullet$ \textbf{Model optimizations}
Various techniques have been proposed to optimize the DL models in consideration of their sizes and computation complexity. Here, we study the usage of two most popular techniques in the wild: quantization and sparsity.
\textbf{Quantization} compresses DL models by reducing the number of bits required to represent each weight. The quantization has different levels, including 16-bit~\cite{gupta2015deep}, 8-bit~\cite{vanhoucke2011improving}, and even 1-bit~\cite{courbariaux2015binaryconnect,rastegari2016xnor}, while the original models are usually presented in 32-bit floating points.
\textbf{Sparsity}~\cite{lebedev2016fast,wen2016learning} has also been extensively studied in prior literatures as an effective approach to make compact DL models.
It's mainly used in matrix multiplication and convolutional layers to reduce parameters.
Such optimizations are long known to reduce DL cost by up to two orders of magnitude without compromising model accuracy~\cite{han2015deep}.

Table~\ref{tab:model_opts} summarizes the optimizations applied on DL models. Here we focus on the DL models for 5 popular frameworks, i.e., \emph{TensorFlow (TF), TFLite, Caffe, Caffe2}, and \emph{ncnn}.
Overall, most of these frameworks only support 8-bit quantization, except \emph{TensorFlow} who has 16-bit quantization and sparsity support.
However, only a fraction of DL models apply the optimization techniques: 6.32\% models are quantized into 8-bit, while others are non-optimized.

\implication{
The findings that well-known DL optimizations are missing in real-world deployment suggest the efficiency potential of mobile DL is still largely untapped. 
The findings also urge immediate actions to fix the missing optimizations. 
}

\subsection{Model Resource Footprint}\label{sec:model_performance}

\begin{figure}[t]
	\centering
	\includegraphics[width=0.48\textwidth]{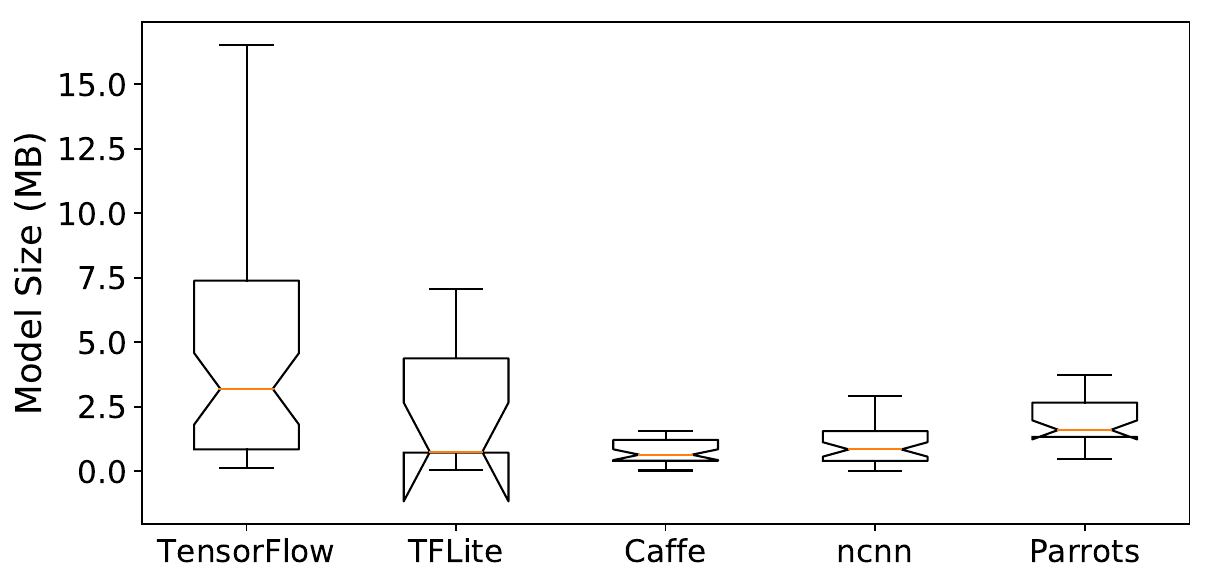}
	\caption{The size of DL models in different frameworks. We leave out \emph{Caffe2} since we only successfully extract one model in \emph{Caffe2} format.}
	\label{fig:model_size}
\end{figure}
\noindent $\bullet$ \textbf{Model size.}
Figure~\ref{fig:model_size} illustrates the size of DL models (in storage). Overall, we find that the extracted DL models are quite small (median: 1.6MB, mean: 2.5MB), compared to classical models such as VGG-16~\cite{VGG} (around 500MB) and MobileNet~\cite{MobileNet} (around 16MB).
The models in \emph{TensorFLow} format (median: 3.2MB) are relatively larger than the models in other formats such as \emph{TFLite} (median: 0.75MB) and \emph{ncnn} (median: 0.86MB).

\begin{figure}[t]
	\centering
	\includegraphics[width=0.48\textwidth]{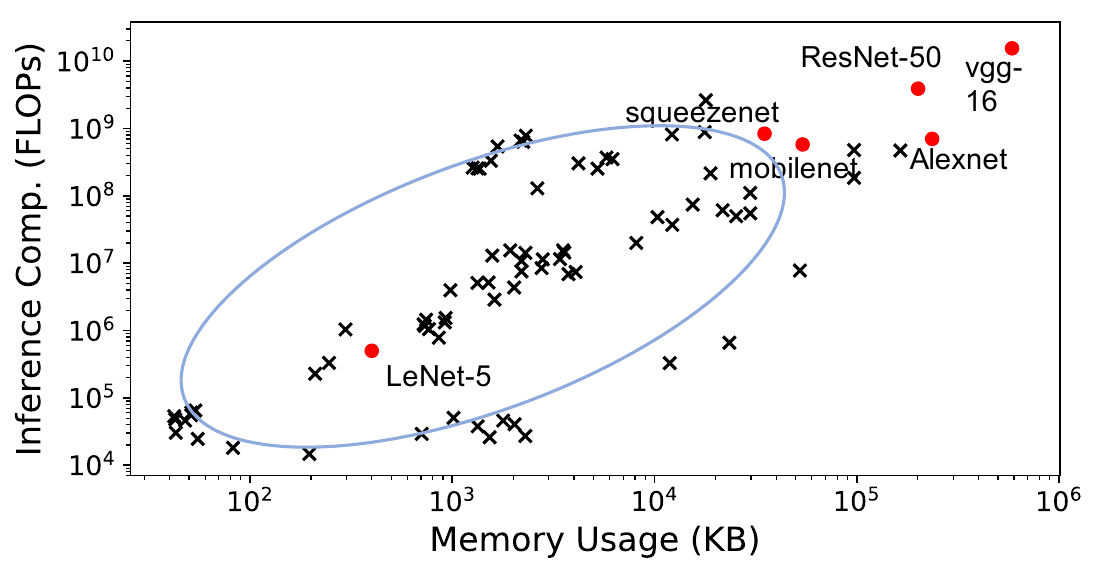}
	\caption{The cost of memory and computation of DL models extracted from apps. Red dots: classical CNN architectures as references. 
	Black crosses: extracted DL models, which are visually summarized by a covariance error ellipse~\cite{ellipse}.
	}
	\label{fig:model_perf}
\end{figure}
\noindent $\bullet$ \textbf{Runtime overhead.}
We then study the runtime performance of DL models. Here, we focus on two aspects: memory and computations. The memory usage includes both the model parameters and the generated intermediate results (feature maps). For computation complexity, we use floating point operations (FLOPs) during one inference as the metric. 
Here we use only part of models in \emph{TensorFlow} and \emph{ncnn} formats since some others don't have fixed input sizes, e.g., image size, so that the computation complexity can only be determined at runtime~\cite{tfshape}. 
We also include the performance of some other classical CNN models such as AlexNet, MobileNet, etc.


As illustrated in Figure~\ref{fig:model_perf}, the black crosses represent the DL models we have extracted, while the red dots represent the classical CNN architectures. Overall, the results show that \textit{in-the-wild DL models are very lightweight in consideration of memory usage and computation complexity, with median value of 2.47 MB and 10M FLOPs respectively.}
Running such models on mobile processors is inexpensive. For example, as estimated on the CPU of Snapdragon 845\footnote{A typical mobile chip used by many popular smartphones such as Galaxy S8.}, the execution time of 80\% models are less than 15ms which is translated to 67 FPS~\cite{snapdragonflops}.
To be compared, ResNet-50, one of the state-of-the-art models in image classification task, has around 200MB memory usage and 4GFLOPs computations. Even MobileNet and SqueezeNet, which are designed and optimized for mobile scenarios, require more memory usage and computations than 90\% those mobile DL models that we have discovered.

\implication{
Our findings of dominant lightweight DL models on smartphones give app developers a valuable assurance: 
DL inference can be as cheap as a few MBs of memory overhead and tens of ms execution delay. 
Our findings challenge existing research on DL inference, which are typically centered on full-blown models (e.g. VGG) and validated on these models~\cite{conf/mobisys/LiuLZNLD18,conf/ipsn/LaneBGFJQK16,conf/asplos/KangHGRMMT17,conf/mobisys/HanSPAWK16,conf/nips/DentonZBLF14,conf/mobisys/LocLB17}.
Given the significance of smartphones as DL platforms, future DL algorithm proposals should consider applicability on lightweight models and treat resource constraints as the first-class concern.}

\subsection{Model Security}\label{sec:model_security}
Finally, we investigate into how DL models are protected. We deem model protection as an important step to AI system/app security, because if attackers can acquire the model, they can (i) steal the intellectual property by reusing the model file or re-train a new model based on the stolen one; (ii) easily attack the DL-based apps via adversarial attack~\cite{papernot2016limitations}.
We focus on two practical protection mechanisms.
\begin{itemize}
\item \textbf{Obfuscation} is a rather shallow approach to prevent attackers from gaining insights into the model structures by removing any meaningful text, e.g., developers-defined layer names.
\item \textbf{Encryption} is better in security by avoiding attackers from getting the model structures/parameters, but also causes inevitable overhead for apps to decrypt the models in memory.
Here, we deem encrypted models as always obfuscated too.
\end{itemize}
We investigate into how obfuscation and encryption are employed on DL models that we have extracted. We analyze the DL models extracted from apps using \emph{TensorFLow, TFLite, ncnn, caffe}, and \emph{Caffe2}. In total, we confirm the security level of 120 DL models. Note that here encryption doesn't necessarily mean encryption algorithm, but also includes cases where developers customize the model format so that the model cannot be parsed via the DL framework.

The results show that among the 120 DL models, we find 47 (39.2\%) models are obfuscated and 23 (19.2\%) models are encrypted.
Note that these two sets of apps are overlapped: encrypted apps are also obfuscated.
The results indicate that \textit{most DL models are exposed without protection, thus can be easily extracted and utilized by attackers.}
In fact, only few frameworks in Table~\ref{tab:overview_of_frameworks} support obfuscation, e.g., \emph{ncnn} can convert models into binaries where text is all striped~\cite{ncnnbin}, and \emph{Mace} can convert a model to C++ code~\cite{macemodel}.
What's worse, no framework provides help in model encryption as far as we know.
Thus, developers have to implement their own encryption/decryption mechanism, which can impose non-trivial programming overhead.

\implication{The grim situation of model security urges strong protection over proprietary models in a way similar to protecting copyright digital contents on smartphones~\cite{nagai2018digital,zhang2018protecting,rouhanideepsigns,adi2018turning}. 
This necessitates a synergy among new tools, OS mechanisms and policies, and hardware mechanisms such as Intel SGX~\cite{sgx} and ARM TrustZone~\cite{trustzone}}.
\section{Limitations and Future Work}\label{sec:limitations}
\textbf{Limitations of our analyzing tool} Though we carefully design our analyzer to capture as many DL apps as possible, and involve a lot of manual efforts to validate the results, we can still have false identifications. For example, those DL apps that neither depend on any popular DL frameworks nor have any string patterns in the native libraries, will be missed out. In addition, the apps that have integrated DL frameworks but don't really use them will be falsely classified as DL apps, which shouldn't happen though.
For the first case, we plan to mine the code pattern of DL implementation and use the pattern to predict more DL apps that might involve DL.
For the second one, we plan to further enhance our analyzer with advanced static analysis technique~\cite{arzt2014flowdroid} so that it can detect whether the API calls (sinks) of DL libraries will be invoked or not.

\textbf{Longer-term analysis} Currently, we carry out our empirical study based on the app datasets obtained in Jun. and Sep. 2018.
In the future, we plan to actively maintain and update our study by extending to more time steps, e.g., every 3 months.
We believe that more solid and interesting results can be made through such long-term analysis.
We will also keep watch on newly emerging DL frameworks and add them to our analyzing tool.

\textbf{More platforms} In this work, we only analyze the adoption of DL on Android apps. Though Android is quite representative of the mobile ecosystem, more interesting findings might be made by expanding our study on other platforms such as iOS and Android Wear.
We believe that comparing the DL adoption on different platforms can feed in more implications to researchers and developers.

\textbf{Involving dynamic analysis} For now, our analysis remains static. Though static analysis technology is quite powerful, we believe that dynamic analysis can provide more useful findings. For example, by running the DL models on off-the-shelf mobile devices, we can characterize the accurate runtime performance, e.g., end-to-end latency and energy consumption.
\section{Related Work}\label{sec:related}
In this section, we discuss existing literature studies that relate to our work in this paper.

\textbf{Mobile DL}
Due to their ubiquitous nature, mobile devices can generate a wide range of unique sensor data, and thus create countless opportunities for DL tasks.
The prior efforts on mobile DL can be mainly summarized into two categories.
First, researchers have built numerous novel applications based on DL~\cite{conf/huc/BarzS16,conf/huc/MittalYGK16,conf/sensys/LaneBGFK15,conf/huc/RaduLBMMK16,conf/huc/LaneGQ15}. For example, MobileDeepPill~\cite{DBLP:conf/mobisys/ZengCZ17} is a small-footprint mobile DL system that can accurately recognize unconstrained pill images.
Second, various optimization techniques have been proposed to reduce the overhead of DL on resource-constrained mobile devices. The optimizations include model compression~\cite{conf/mobisys/LiuLZNLD18,conf/ipsn/LaneBGFJQK16,conf/cvpr/WuLWHC16,conf/nips/DentonZBLF14,conf/mobisys/HanSPAWK16}, hardware customizations~\cite{conf/asplos/ChenDSWWCT14,conf/fpga/ZhangLSGXC15,conf/isca/ChenES16,conf/isca/HanLMPPHD16}, lightweight models~\cite{conf/huc/LaneGQ15,conf/icassp/ChenPH14,conf/icassp/VarianiLMMG14,MobileNet}, knowledge distilling~\cite{hinton2015distilling,balan2015bayesian,DBLP:conf/mobisys/ZengCZ17}, and cloud offloading~\cite{conf/asplos/KangHGRMMT17,conf/isca/HauswaldKLCLMDM15,journals/tc/ZhangYC16}.
These studies are usually carried out under lab environments, based on classical models such as VGG and ResNet, in lack of real-world workloads and insights.
Thus, our work is motivated by those enormous efforts that try to bring DL to mobile devices, and fill the gap between the academic literature and industry products.

\textbf{ML/DL as cloud services}
Besides on-device fashion, the DL functionality, or in a broader scope of Machine Learning (ML), can also be accessed as cloud services.
The service providers include Amazon~\cite{amazonml}, Google~\cite{googleml}, Microsoft Azure~\cite{azureml}, etc.
There are some prior analyzing studies focusing on such MLaaS (machine learning as a service) platforms.
Yao \textit{et al.}~\cite{conf/imc/YaoXWVZZ17} comprehensively investigate into effectiveness of popular MLaas systems, and find that with more user control comes greater risk.
Some other literature studies~\cite{fredrikson2015model,shokri2017membership,tramer2016stealing} focus on the security issues of those platforms towards different types of attacks.
MLaaS platforms have some advantages in protecting intellectual property and performance. However, compared to on-device fashion, they also have some shortcomings such as privacy concerns and unstable accessibility.
Thus, some DL tasks are more fit to be run on local devices, e.g., word-prediction in keyboard.
Our work also proves that on-device DL has been already adopted in many real-world apps to some extent, and is going to be popular on smartphones.

\textbf{Empirical study of DL}
Prior empirical analysis mainly focuses on assisting developers to build better DL apps/systems/models.
Zhang \textit{et al.}~\cite{zhang2018empirical} characterize the defects (bugs) in DL programs via mining the StackOverflow QA pages and GitHub projects.
Consequently, the results are limited in only open-source, small-scale projects.
Fawzi \textit{et al.}~\cite{fawzi2018empirical} analyze the topology and geometry of the state-of-the-art deep networks, as well as their associated decision boundary.
Senior \textit{et al.}~\cite{senior2013empirical} investigate into how the learning rate used in stochastic gradient descent impacts the training performance, and propose schemes to select proper learning rate.
These studies mostly focus on classical and small-scale DL models proposed in previous literature, while our study mine the knowledge from large-scale in-the-wild mobile apps.

\textbf{DL Model protection}
Some recent efforts have been investigated in protecting DL models. For example, various watermarking mechanisms~\cite{nagai2018digital,zhang2018protecting,rouhanideepsigns,adi2018turning} have been proposed to protect intellectual property of DL models. This approach, however, cannot protect models from being extracted and attacked.
As a closer step, some researchers~\cite{tramer2018slalom,hanzlik2018mlcapsule,gu2018securing} secure the DL systems/models based on secure execution environments (SEE) such as Intel SGX~\cite{sgx}.
However, those techniques are still not practical for in-the-wild deployment.
Some DL frameworks also provide mechanisms for model protection. For example, \emph{Mace}~\cite{mace} supports developers to convert models to C++ code~\cite{macemodel}.
However, our results show that a large number of DL models are exposed without secure protection.
\section{Conclusions}\label{sec:conclusion}
In this work, we have carried out the first empirical study to understand how deep learning technique is adopted in real-world smartphones, as a bridge between the research and practice.
By mining and analyzing large-scale Android apps based on a static tool, we have reached interesting and valuable findings.
For example, we show that early adopters of mobile deep learning are top apps, and the role played by deep learning in those apps is critical and core.
Our findings also provide strong and valuable implications to multiple stakeholders of the mobile ecosystem, including developers, hardware designers and researchers.

\bibliographystyle{ACM-Reference-Format}
\bibliography{ref}

\nocite{liu2007towards}

\end{document}